\documentclass[runningheads]{llncs}
\usepackage{makeidx}
\usepackage{graphicx}
\usepackage{amsmath,amssymb} 
\usepackage{color}

\usepackage{tikz, pgfplots}
\usetikzlibrary{calc}
\usetikzlibrary{positioning}
\usepackage{verbatim}
\tikzset{basic/.style={draw,fill=blue!20,text width=1em,text badly centered}}

\tikzset{input/.style={basic,circle}}
\tikzset{weights/.style={basic,rectangle}}
\tikzset{functions/.style={basic,circle,fill=blue!10}}
\tikzset{activation/.style={basic,circle}}
\usepackage{subfigure}
\usepackage{booktabs}
\usepackage{cite}
\usepackage[font={small},skip=0pt]{caption}
\usepackage{tabularx}

\begin{document}
	\pagestyle{headings}
	\mainmatter

	\def\GCPR17SubNumber{58}

	\title{Recurrent Residual Learning for Action Recognition}

	\titlerunning{Recurrent Residual Learning for Action Recognition}
	\authorrunning{Ahsan Iqbal et al.}
	\author{Ahsan Iqbal, Alexander Richard, Hilde Kuehne, Juergen Gall \\
	        \texttt{\{iqbalm,richard,kuehne,gall\}@iai.uni-bonn.de}}
	\institute {University of Bonn, Germany}

	\maketitle

	\begin{abstract}
	Action recognition is a fundamental problem in computer vision with a lot of potential applications such as video surveillance, human computer interaction, and robot learning. Given pre-segmented videos, the task is to recognize actions happening within videos. Historically, hand crafted video features were used to address the task of action recognition. With the success of Deep ConvNets as an image analysis method, a lot of extensions of standard ConvNets were purposed to process variable length video data. In this work, we propose a novel recurrent ConvNet architecture called recurrent residual networks to address the task of action recognition.
The approach extends ResNet, a state of the art model for image classification. While the original formulation of ResNet aims at learning spatial residuals in its layers, we extend the approach by introducing recurrent connections that allow to learn a spatio-temporal residual. In contrast to fully recurrent networks, our temporal connections only allow a limited range of preceding frames to contribute to the output for the current frame, enabling efficient training and inference as well as limiting the temporal context to a reasonable local range around each frame. On a large-scale action recognition dataset, we show that our model improves over both, the standard ResNet architecture and a ResNet extended by a fully recurrent layer.
	\end{abstract}

	\section{Introduction}			
	\label{sec:introduction}
Action recognition in videos is an important research topic \cite{NIPS2014_5353, NIPS2016_6433, Bilen16a} with many potential applications such as video surveillance, human computer interaction, and robotics. Traditionally, action recognition has been addressed by hand crafted video features in combination with classifiers like SVMs as in \cite{HengWang:2011:ARD:2191740.2192078, wang:hal-00873267}. With the impressive achievements of deep convolutional networks (ConvNets) for image classification, a lot of research was devoted to extend ConvNets to process video data, however, with unsatisfying results. While ConvNets have shown to perform very well for spatial data, they perform poorly for temporal data since they fail to model temporal dependencies. Heuristics were therefore developed for modeling temporal relations. First attempts, which simply stacked the frames and used a standard ConvNet for image classification \cite{Karpathy_2014_CVPR}, performed worse than hand crafted features. More successful have been two stream architectures \cite{NIPS2014_5353} that use two ConvNets. While the first network is applied to the independent frames, the second network processes the optical flow, which needs to be computed beforehand. While two stream architectures achieve lower classification error rates than hand-crafted features, they are very expensive for training and inference since they need two ConvNets and an additional approach to extract the optical flow.        

In this work, we propose a more principled way to integrate temporal dependencies within a ConvNet. Our model is based on the state of the art residual learning framework \cite{resnet} for image classification, which learns a residual function with respect to the layer's input. We extend the approach to a sequence of images by having a residual network for each image and connecting them by recurrent connections that model temporal residuals. In contrast to the two stream architecture  \cite{NIPS2016_6433}, which proposes residual connections from the motion to the appearance stream, our approach is a single stream architecture that directly models temporal relations within the spatial stream and does not require the additional computation of the optical flow.    
    
We evaluate our approach on the popular UCF-101 \cite{DBLP:journals/corr/abs-1212-0402} benchmark and show that our approach reduces the error of the baseline \cite{resnet} by 17\%. Although two stream architectures, which require the computation of the optical flow, achieve a lower error rate, the proposed approach of temporal residuals could also be integrated into a two stream architecture.        		

	\section{Related Work}			
	\label{sec:relatedwork}
  
Due to the difficulty of modeling temporal context with deep neural networks, traditional methods using hand-crafted features have been state of the art in action recognition much longer than in image classification~\cite{laptev2005space,HengWang:2011:ARD:2191740.2192078,wang:hal-00873267,wang2016mofap}.
The most popular approaches are dense trajectories~\cite{HengWang:2011:ARD:2191740.2192078} with a bag-of-words and SVM classification as well as improved dense trajectories~\cite{wang:hal-00873267} with Fisher vector encoding.
Due to the success of deep architectures, first attempts in action recognition aimed at combining those traditional features with deep models. In~\cite{richard2017bow}, for instance, a combination of hand crafted features and recurrent neural networks have been deployed. Peng et.\ al.~\cite{Peng2014} proposed Stacked Fisher Vectors, a video representation with multi-layer nested Fisher vector encoding. In the first layer, they densely sample large subvolumes from input videos, extract local features, and encode them using Fisher vectors. The second layer compresses the Fisher vectors of subvolumes obtained in the previous layer, and then encodes them again with Fisher vectors. Compared with standard Fisher vectors, stacked Fisher vectors allow to refine and abstract semantic information in a hierarchical way. Another hierarchical approach has been proposed in \cite{Jhuang:ICCV:207}, who apply HMAX \cite{riesenhuber1999hierarchical} with pre-defined spatio-temporal filters in the first layer.
Trajectory pooled deep convolutional descriptors are defined in \cite{WangQT15a}. CNN features are extracted from a two stream architecture and are combined with improved dense trajectories.

In the past, there have been attempts to address the task of action recognition with deep architectures directly. However, in most of these works, the input to the model is a stack of consecutive video frames and the model is expected to learn spatio-temporal dependent features in the first few layers, which is a difficult task. In \cite{Taylor:2010:CLS:1888212.1888225, Chen:2010, DBLP:conf/cvpr/LeZYN11}, spatio temporal features are learned in unsupervised fashion by using Restricted Boltzmann machines.
The approach of \cite{Jain_2015_CVPR} combines the information about objects present in the video with the motion in the videos. 3D convolution is used in \cite{Ji:2013:CNN:2412386.2412939} to extract discriminative spatio temporal features from the stack of video frames. Three different approaches (early fusion, late fusion, and slow fusion) were evaluated to fuse temporal context in \cite{Karpathy_2014_CVPR}. A similar technique as in \cite{Ji:2013:CNN:2412386.2412939} is used to fuse temporal context early in the network, in late fusion, individual features per frame are extracted and fused in the last convolutional layer. Slow fusion mixes late and early fusion. In contrast to these methods, our method does not rely on temporal convolution but on a recurrent network architecture directly.

More recently, \cite{Bilen16a} proposed concept of dynamic images. The dynamic image is based on the rank pooling concept \cite{DBLP:journals/corr/FernandoGMGT15} and is obtained through the parameters of a ranking machine that encodes the temporal evolution of the frames of the video. Dynamic images are obtained by directly applying rank pooling on the raw image pixels of a video producing a single RGB image per video. And finally, by feeding the dynamic image to any CNN architecture for image analysis, it can be used to classify actions.

The most successful approach to date is the two-stream CNN of \cite{NIPS2014_5353}, where individual frames from the videos are the input to the spatial network, while motion in the form of dense optical flow is the input to the temporal network. The features learned by both networks are concatenated and finally linear SVM is used for classification. 
Recently, with the success of ResNet \cite{resnet}, \cite{NIPS2016_6433} proposed a model that combines ResNet and the two stream architecture. They replace both spatial and temporal networks in the two stream architecture by a ResNet with 50 layers. They also introduce a temporal or motion residual, i.e.\ a residual connection from the temporal network to the spatial network to enable learning of spatio temporal features. In contrast to our method, they incorporate temporal information by extending the convolutions over temporal windows. Note that this leads to a largely increased amount of model parameters, whereas our approach shares the weights among all frames, keeping the network size small. 
\cite{WangXWQLTV16} proposed the temporal segment networks, which are mainly based on the two stream architecture. However, rather than densely sampling  every other frame in the video, they divide the video in segments of equal length, and then randomly sample snippets from these segments as network input. In this way, the two stream network produces segment level classification scores, which are combined to produce video level output.

Deep recurrent CNN architectures are also explored to model dependencies across the frames. In \cite{DBLP:journals/corr/DonahueHGRVSD14}, convolutional features are fed into an LSTM network to model temporal dependencies. \cite{Yang:2016:MMF:2964284.2964297} considered four networks to address action recognition in videos. The first network is similar to spatial network in the two stream architecture. The second network is a CNN with one recurrent layer, it expects a single optical flow image and in recurrent layer, optical flows over a range of frames are combined. In the third network, they feed a stack of consecutive frames, the network is also equipped with a recurrent layer to capture the long term dependencies. Similarly, the fourth network expects a stack of optical flow fields as input. However, the network is equipped with a fully connected recurrent layer. Finally, boosting is used to combine the output of all four networks.

Finally,~\cite{wang2016recurrent} equip a ResNet with recurrent skip connections that are, contrary to ours, purely temporal skip connections, whereas in our framework, we use spatio-temporal skip connections. Note the significant difference in both approaches: while purely temporal skip connections can be interpreted as usual recurrent connections with unit weights, spatio-temporal skip connections are a novel concept that allow for efficient backpropagation and combine both, changes in the temporal domain and changes in the spation domain at the same time.

\section{Recurrent Residual Network}
\label{sec:spatio_temporal_residual_network}
In this section, we describe our approach to address the problem of action recognition in videos. Our approach is an extension of ResNet \cite{resnet}, which reformulates a layer as learning the spatial residual function with respect to the layer's input. State of the art results were achieved in image recognition tasks by learning spatial residual functions. We extend the approach to learn temporal residual functions across the frames to do action recognition in videos. In our formulation, the feature vector at time step $t$ is a residual function with respect to the feature vector at time step $t-1$. By following the analogy of ResNet, temporal residuals are learned by introducing the temporal skip (recurrent) connections. In the following, we give a brief introduction to ResNet, explain different types of temporal skip connections, and finally describe how to include more temporal context.

\subsection{ResNet}
\begin{figure}
\centering
\subfigure[Overall ResNet scheme] {
\begin{tikzpicture}[node distance=1cm]
\tikzstyle{io} = [rectangle, rounded corners, minimum width=1cm, minimum height=0.5cm,text centered, draw=black]
\tikzstyle{layer} = [rectangle, rounded corners, minimum width=2cm, minimum height=0.5cm,text centered, draw=black]
\tikzstyle{arrow} = [thick,->,>=stealth]

\draw [color=white] (-2.0, 0.0) -- (2.0, 0.0); 

\node (x) [io] {$x$};
\node (c1) [layer, below of=x] {Init. Layers};
\node (b1) [layer, below of=c1] {Block 1};
\node (b2) [layer, below of=b1] {Block 2};
\node (b3) [layer, below of=b2] {Block 3};
\node (b4) [layer, below of=b3] {Block 4}; 
\node (p2) [layer, below of=b4] {Pool};
\node (fc) [io, below of=p2] {softmax};

\draw [arrow] (x) to (c1);
\draw [arrow] (c1) to (b1);
\draw [arrow] (b1) to (b2);
\draw [arrow] (b2) to (b3);
\draw [arrow] (b3) to (b4);
\draw [arrow] (b4) to (p2);
\draw [arrow] (p2) to (fc);

\end{tikzpicture}
}
\hfill
\subfigure[A single ResNet block] {
\begin{tikzpicture}[node distance=1cm]
\tikzstyle{io} = [rectangle, rounded corners, minimum width=1cm, minimum height=0.5cm,text centered, draw=black]
\tikzstyle{layer} = [rectangle, rounded corners, minimum width=1.5cm, minimum height=0.5cm,text centered, draw=black]
\tikzstyle{arrow} = [thick,->,>=stealth]
\tikzstyle{dashedarrow} = [dashed,->,>=stealth]
\tikzstyle{line} = [thick,-]

\node (l0) [io] {$x$};
\node (t0) [left=0.1cm of l0] {};
\node (l1) [layer, below of=l0] {layers};
\node (l3) [io, below of=l1] {$+$};
\node (l4) [layer, below of=l3] {layers};
\node (l5) [io, below of=l4] {$+$};
\node (l6) [layer, below of=l5] {layers};
\node (l7) [io, below of=l6] {$+$};
\node (l8) [below of=l7] {};
\node (t8) [left=0.1cm of l7] {};

\draw [arrow] (l0) to (l1);
\draw [arrow] (l1) to (l3);
\draw [arrow] (l0) to [out=360,in=360,looseness=1.0] (l3);
\draw [arrow] (l3) to (l4);
\draw [arrow] (l3) to [out=360,in=360,looseness=1.0] (l5);
\draw [arrow] (l4) to (l5);
\draw [arrow] (l5) to (l6);
\draw [dashedarrow] (l5) to [out=360,in=360,looseness=1.0] node[right] {Downsampling} (l7);
\draw [arrow] (l6) to (l7);
\draw [arrow] (l7) to (l8);

\draw [line] (t0) to [out=180,in=180,looseness=0.5] node[left] {Block x} (t8);
\end{tikzpicture}
}
\caption{ResNet architecture, (a) shows the overall ResNet structure with four building blocks and a final classification layer, (b) is the schema of a single block: each block consists of multiple convolutional layers and skip connections to learn the residuals. At the end, the output feature maps are downsampled.}
\label{fig_resnet}
\end{figure}
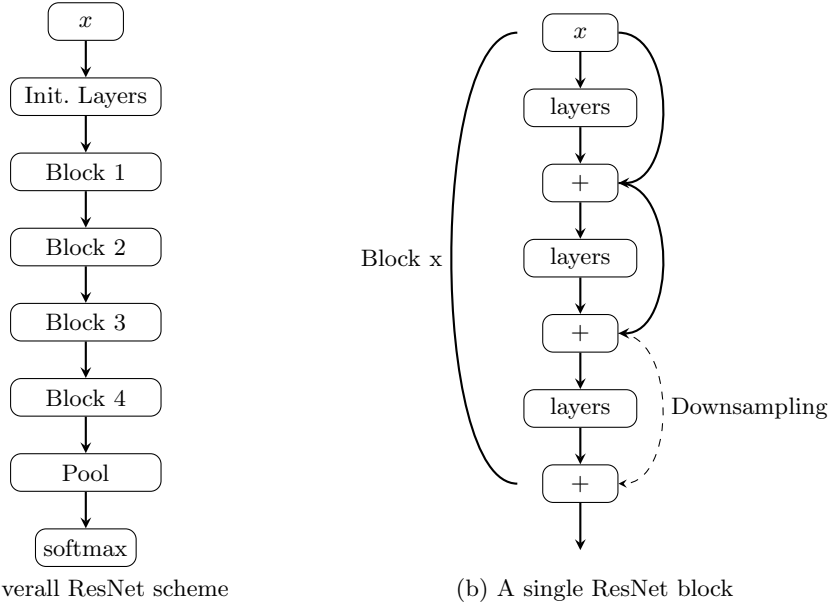

ResNet \cite{resnet} introduces a residual learning framework. In this framework, a stack of convolutional layers fit a residual mapping instead of the desired mapping. Let $H(x)$ denote the desired mapping.
The principle of ResNet is to interpret the mapping of the learned function from one layer to another as $ H(x) = F(x) + x $, i.e.\ as the original input $ x $ plus a residual function $ F(x) $. Introducing the spatial skip connection, the input signal $ x $ is directly forwarded and added to the next layer, so it only remains to learn the residual $ F(x) = H(x) - x $, see Figure~\ref{fig_resnet}b.

\subsection{Type of Temporal Skip Connection}

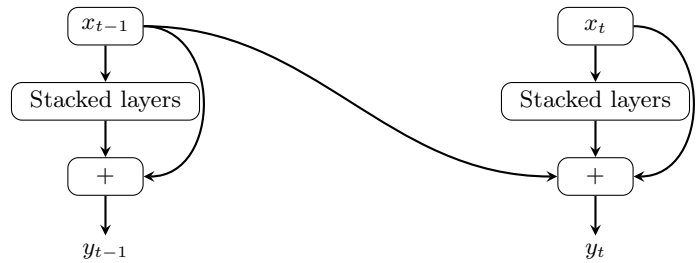
\begin{figure}
\centering
\begin{tikzpicture}[node distance=1cm]
\tikzstyle{io} = [rectangle, rounded corners, minimum width=1cm, minimum height=0.5cm,text centered, draw=black]
\tikzstyle{layer} = [rectangle, rounded corners, minimum width=2.5cm, minimum height=0.5cm,text centered, draw=black]
\tikzstyle{arrow} = [thick,->,>=stealth]

\node (l0) [io] {$x_{t-1}$};
\node (l1) [layer, below of=l0] {Stacked layers};
\node (l3) [io, below of=l1] {$+$};
\node (l4) [below of=l3] {$y_{t-1}$};

\draw [arrow] (l0) to (l1);
\draw [arrow] (l1) to (l3);
\draw [arrow] (l0) to [out=360,in=360,looseness=1.35] (l3);
\draw [arrow] (l3) to (l4);

\node (m0) [io, right=5.5cm of l0] {$x_t$};
\node (m1) [layer, below of=m0] {Stacked layers};
\node (m3) [io, below of=m1] {$+$};
\node (m4) [below of=m3] {$y_{t}$};

\draw [arrow] (m0) to (m1);
\draw [arrow] (m1) to (m3);
\draw [arrow] (m0) to [out=360,in=360,looseness=1.35] (m3);
\draw [arrow] (m3) to  (m4);

\draw [arrow] (l0) to [out=360, in=180] (m3);
\end{tikzpicture}
\caption{Building block of our recurrent residual learning, $x_{t-1}$ represents the input to a convolutional layer at time step $t-1$ and $x_t$ represents the input to the same layer at time step $t$. While the spatial skip connections within a single time frame allow to learn a spatial residual, the spatio-temporal skip connection from time $ t-1 $ to time $ t $ adds temporal information to the learned residual. }
\label{fig_sp_tp_resnet}
\end{figure} 

There are multiple possibilities to model the temporal skip connection. The standard spatial skip connections in the classical ResNet architecture are either an identity mapping, i.e.\ they just forward the input signal and add it to the destination layer, or they perform a linear transformation in order to establish the downsampling as depicted in Figure~\ref{fig_resnet}.
The simplest case for the temporal skip connection is to also use an identity mapping. With the notation of Figure~\ref{fig_sp_tp_resnet}, the layer output $ y_t $ at time $ t $ is the residual function
\begin{align}
    y_t = \sigma(x_t * W) + x_t,
\end{align}
where $ \sigma $ represents the nonlinear operations performed after the linear transformation. Note that for simplicity of notation, we pretend that the residual block contains a single convolutional layer only and $W$ represents weights for the layer. Extending this for the temporal skip connection, we obtain
\begin{align}
    y_t = \sigma(x_t * W) + x_t + x_{t-1}.
    \label{eq_identity_mapping}
\end{align}
In order to allow for a weighting of the temporal skip connection with weights $W_s$, a linear transformation can be applied to $ x_{t-1} $ before adding it to $ y_t $,
\begin{align}
    y_t = \sigma(x_t * W) + x_t + x_{t-1} * W_s.
    \label{eq_linear_mapping}
\end{align}
Moreover, in order to learn a nonlinear spatio-temporal mapping, this can be further extended to
\begin{align}
    y_t = \sigma(x_t * W) + x_t + \sigma(x_{t-1} * W_s).
    \label{eq_nlinear_mapping}
\end{align}

\subsection{Temporal Context}

\begin{figure}
\centering
\begin{tikzpicture}[node distance=1cm]
\tikzstyle{io} = [rectangle, rounded corners, minimum width=1.0cm, minimum height=0.5cm,text centered, draw=black]
\tikzstyle{layer} = [rectangle, rounded corners, minimum width=2.5cm, minimum height=0.5cm,text centered, draw=black]
\tikzstyle{arrow} = [thick,->,>=stealth]

\node (x_2)  {$x_{t-2}$}; 
\node (ndots) [below of=x_2] {$\vdots$} -- (ndots);
\node (n0) [io, below of=ndots] {ReLu};
\node (n1) [layer, below of=n0] {Stacked layers};
\node (n3) [io, below of=n1] {$+$};
\node (n4) [io, below of=n3] {ReLu};

\node (n5) [layer, below of=n4] {Stacked layers};
\node (n7) [io, below of=n5] {$+$};
\node (n8) [below of=n7] {$y_{t-2}$};

\draw [arrow] (x_2) to (ndots);
\draw [arrow] (ndots) to (n0);
\draw [arrow] (n0) to (n1);
\draw [arrow] (n1) to (n3);
\draw [arrow] (n0) to [out=360,in=360,looseness=1.35] (n3);

\draw [arrow] (n3) to (n4);
\draw [arrow] (n4) to (n5);
\draw [arrow] (n5) to (n7);
\draw [arrow] (n7) to (n8);

\draw [arrow] (n4) to [out=360,in=360,looseness=1.35] (n7);

\node (x_1) [right=3.5cm of x_2] {$x_{t-1}$}; 
\node (odots) [below of=x_1] {$\vdots$} -- (odots);
\node (o0) [io, below of=odots] {ReLu};
\node (o1) [layer, below of=o0] {Stacked layers};
\node (o3) [io, below of=o1] {$+$};
\node (o4) [io, below of=o3] {ReLu};

\node (o5) [layer, below of=o4] {Stacked layers};
\node (o7) [io, below of=o5] {$+$};
\node (o8) [below of=o7] {$y_{t-1}$};

\draw [arrow] (x_1) to (odots);
\draw [arrow] (odots) to (o0);
\draw [arrow] (o0) to (o1);
\draw [arrow] (o1) to (o3);
\draw [arrow] (o0) to [out=360,in=360,looseness=1.35] (o3);

\draw [arrow] (o3) to (o4);
\draw [arrow] (o4) to (o5);
\draw [arrow] (o5) to (o7);
\draw [arrow] (o7) to (o8);

\draw [arrow] (o4) to [out=360,in=360,looseness=1.35] (o7);

\node (x) [right=3.5cm of x_1] {$x_{t}$}; 
\node (pdots) [below of=x] {$\vdots$} -- (pdots);
\node (p0) [io, below of=pdots] {ReLu};
\node (p1) [layer, below of=p0] {Stacked layers};
\node (p3) [io, below of=p1] {$+$};
\node (p4) [io, below of=p3] {ReLu};

\node (p5) [layer, below of=p4] {Stacked layers};
\node (p7) [io, below of=p5] {$+$};
\node (p8) [below of=p7] {$y_{t}$};

\draw [arrow] (x) to (pdots);
\draw [arrow] (pdots) to (p0);
\draw [arrow] (p0) to (p1);
\draw [arrow] (p1) to (p3);
\draw [arrow] (p0) to [out=360,in=360,looseness=1.35] (p3);

\draw [arrow] (p3) to (p4);
\draw [arrow] (p4) to (p5);
\draw [arrow] (p5) to (p7);
\draw [arrow] (p7) to (p8);
\draw [arrow] (p4) to [out=360,in=360,looseness=1.35] (p7);

\draw [arrow] (n0) to [out=360, in=180] (o3);
\draw [arrow] (n4) to [out=360, in=180] (o7);

\draw [arrow] (o0) to [out=360, in=180] (p3);
\draw [arrow] (o4) to [out=360, in=180] (p7);
\end{tikzpicture}
\caption{A network with two temporal skip connections, capable of handling temporal context of three time steps, omitted layers are normal ResNet blocks, i.e.\ without any temporal skip connections.}
\label{fig_temp_context}
\end{figure}
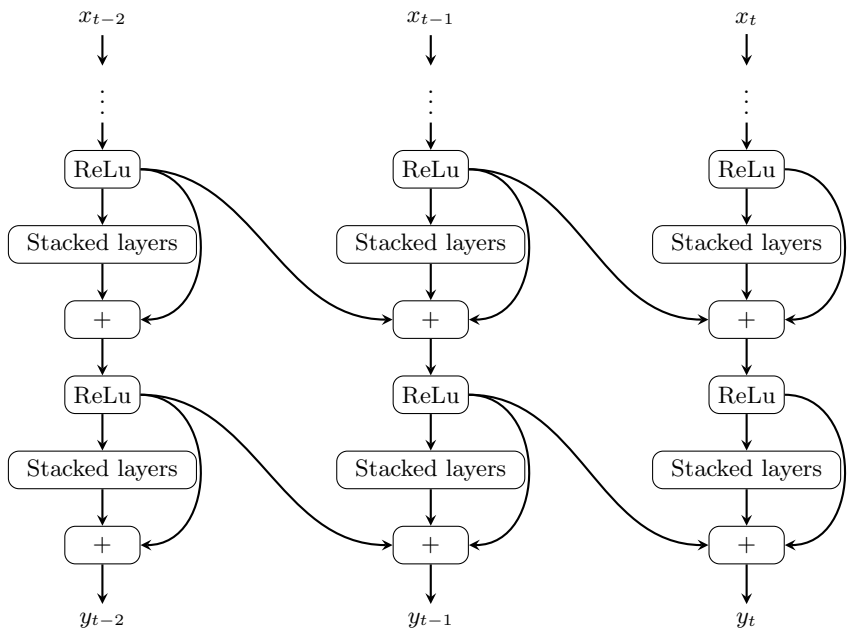

While recurrent connections in traditional recurrent neural networks feed the output of a layer at time $ t-1 $ to the same layer as input at time $ t $, our proposed spatio-temporal skip connections are different. For an illustration, see Figure~\ref{fig_temp_context}. Here, we unfold a network with two spatio-temporal skip connections over time. Note that the temporal context that influences the output $ y_t $ includes $ x_{t-2} $, $ x_{t-1} $, and $ x_t $ as there are paths from $ y_t $ leading to all these inputs. If we only used one temporal skip connection instead of two, the accessible temporal context for $ y_t $ would only be $ x_{t-1} $ and $ x_t $, respectively. In general, if a temporal context over $ T $ frames is desired, at least $ T-1 $ temporal skip connections are necessary.

In order to use this approach for action recognition, a video is divided into $M$ small sequences each containing $T$ frames. A recurrent residual network with $T-1$ temporal skip connections is created to capture the dependencies over these $T$ time steps. In training, we optimize the cross-entropy loss of each small video chunk.
During inference, for each small sequence $\{x_t^{(i)}\}_{t=1}^T$ within one video, the recurrent residual network computes $P(y=c|\{x_t^{(i)}\}_{t=1}^T)$. In order to obtain an overall classification of a complete video, the individual output probabilities are averaged over the $M$ subsequences of the video. Note that this is similar to existing frame-wise approaches, where an output probability per frame is computed and the overall video action probabilities are obtained by accumulating all single frame probabilities. In our case, instead of frames, we use small subsequences of the original video.

\section{Experimental Setup}
\label{sec:experiments}
In this section, we describe our experimental setup. We use ADAM \cite{adam} as learning algorithm and except for the baseline experiments, we update the model after observing 1\% of training data, and every tenth frame from each video is sampled as input to the model. We evaluate our approach on UCF-101 \cite{DBLP:journals/corr/abs-1212-0402}, a large-scale action recognition dataset consisting of 13,000 videos from $ 101 $ different classes. The dataset comprises about $ 2.5 $ million frames in total. All the experimental work was done using our framework squirrel \footnote{https://github.com/alexanderrichard/squirrel}. In the following, we describe the baseline experiments and the experiments with our proposed recurrent residual network.

\subsection{Baseline Experiments}
As a baseline, we extract imagenet \cite{imagenet} features for individual frames in the video. Averaged individual feature vectors represent the feature vector of the complete video. Feature vectors for individual frames  are extracted using a pre-trained ResNet model with 50 layers. A batch normalization layer is added after each layer to normalize the input to a layer. This way the network has in total 106 layers.
\par
We extract the imagenet feature vectors for each frame at three different positions of ResNet, i.e.\ after block4, block3, and block2 respectively, see Figure \ref{fig_resnet}. 
We performed two sets of experiments on extracted features for each block. In one set, we average the frame level feature vectors, after Z-normalization and without Z-normalization, and train a linear classifier. We call this model the average pooling model. Similarly, in the other set, we use a recurrent neural network with 128 gated recurrent units (GRUs) in order to evaluate the performance of a classical recurrent network. We call this model the GRU.

\begin{table}[t]
\centering
\begin{tabularx}{\textwidth}{XXrr}
\toprule
Features & Method & Error Rate (with Z-Norm) & Error Rate (without Z-Norm) \\
\hline
Block4 & Avg. Pool & \textbf{0.236} & 0.237 \\
Block4 & GRU & 0.239 & 0.276 \\
\hline 
Block3 & Avg. Pool & 0.309 & 0.313 \\
Block3 & GRU & 0.403 & 0.325\\
\hline
Block2 & Avg. Pool & 0.431 & 0.434\\
Block2 & GRU & 0.440 & 0.493\\
\hline
\end{tabularx}
\caption{Results of the baseline experiments.}
\label{table:baseline_imagenet}
\vspace{-0.7cm}
\end{table}
Table \ref{table:baseline_imagenet} shows the baseline experiments with imagenet features. The average pooling model outperforms the model with gated recurrent units. Also, it is evident from the experiments that with more depth, features become richer. Hence, the depth plays a significant role in getting good classification accuracy.

\subsection{Effect of type and position of the recurrent connection}

In this set of experiments, we evaluate different types of temporal skip connections along with their position in our proposed model. We evaluate temporal skip connections at four different positions, i.e.\ at the beginning by making the first skip connection in block1 recurrent (referred to as Block1), in the middle by making last skip connection in block2 recurrent (referred to as Block2), by making last convolutional skip connection in block4 recurrent (referred to as Mid Block4), and finally by making last skip connection in block 4 recurrent (referred to as Block4). Also, we evaluate the type of recurrent connections. In these experiments, we evaluate identity mapping temporal skip connections, and temporal skip connections with convolutional weights having kernels of size $1\times1$.
Table \ref{table:exp_position} shows the deeper we place the temporal skip connection in the network, the better is the classification accuracy.
\par
In another set of experiments, we evaluate the effect of the type of temporal skip connection. We change the configuration of the best working setup, i.e.\ the one with the skip connection in block4. The connection performs a parametrized linear or non linear transformation and identity mapping.

\begin{table}[t]
\centering
\small
\begin{tabularx}{0.8\textwidth}{XXr}
\toprule
Position & Type & Error Rate \\
\hline
Block1 & Convolutional & 0.265 \\
Block2 & Convolutional & 0.234\\
Mid Block4 & Convolutional & 0.231\\
Block4 & Convolutional & \textbf{0.219} \\
\hline
\end{tabularx}
\caption{Placing the recurrent connection at different positions in the network.}
\label{table:exp_position}
\vspace{-0.3cm}
\end{table}
Table \ref{table:exp_type} shows the results achieved by different type of connections, placed closer to the output layer as our previous analysis shows that works best. Identity mapping connection with non trainable weights performed best, possibly because introducing more weights in the network causes overfitting. 

\subsection{Effect of Temporal Context}

In this set of experiments, we explore the effect of temporal context. As discussed earlier, with more recurrent connections, the network is able to include additional temporal dependencies. We already investigated the network with one recurrent connection that is able to include temporal context of two frames. In these experiments, we further explore the temporal context of three frames (by introducing two temporal connections in the network), and the temporal context of five frames (by introducing four temporal skip connections in the network). Figure \ref{fig_temp_context} shows the network architecture to accommodate temporal context of three frames.

\begin{table}[t]
\centering
\small
\begin{tabularx}{0.8\textwidth}{Xr}
\toprule
Type & Error Rate \\
\hline
Identity Mapping & \textbf{0.197} \\
Conv. Linear & 0.219\\
Conv. Non-Linear & 0.210\\
\hline
\end{tabularx}
\caption{Results achieved by different type of recurrent connections.}
\label{table:exp_type}
\vspace{-0.7cm}
\end{table}
As it is evident in Table \ref{table:exp_more_temp}, we do not gain much by including more temporal context. The accuracy improves in case of temporal context three, however it gets worse in case of temporal context five. Hence, considering training time, we consider the model with only one temporal skip connection as the best model. Note that due to the fact that we sample every tenth frame from the video, the overall temporal range is actually ten frames. More precisely, the network learns spatio-temporal residuals between the frames $ x_t $ and $ x_{t-10} $, covering a reasonable amount of local temporal progress within the video.
   
\begin{table}[t]
\centering
\begin{tabularx}{0.8\textwidth}{XXr}
\toprule
Temporal Context & Model & Error Rate \\
\hline
1 & baseline & 0.236 \\
2 & 1 recurrent connection & 0.197 \\
3 & 2 recurrent connections & \textbf{0.194} \\
5 & 4 recurrent connections & 0.209 \\ 
\hline
\end{tabularx}
\caption{Results achieved by including more temporal context. For the best setup (context two), the error is reduced by 17\% from 0.236 to 0.194.}
\label{table:exp_more_temp}
\end{table}

We further evaluate our best model on all three splits of UCF-101. On average our best model achieves \textbf{0.198} on UCF-101 \cite{DBLP:journals/corr/abs-1212-0402}, which is a relative improvement of $ 17\% $ over the ResNet baseline which has an error rate of $ 0.236 $.

\subsection{Comparison with the state of the art}

In this section, our best model (with one temporal skip connection and with sample rate 10) is compared with state of the art action recognition methods. As motion in the frames and appearance in individual frames are two complementary aspects for action recognition, most of the state of the art methods consider two different neural networks, an appearance stream and a motion stream, to extract and use appearance and motion for action recognition. The output of both the networks is fused, and a simple classifier is trained to classify videos. As our model uses the raw video frames only rather than optical flow fields,  fair comparison of our model and the state of the art is only possible for the appearance stream. For completeness, we also compare our results with results achieved after the outputs of the appearance and the motion streams are fused, see Table~\ref{table_compare_state_of_the_art}.
\begin{table}[t]
\centering
\small
\begin{tabularx}{\textwidth}{Xrlrlr}
\toprule
Method & Appearance & \phantom{ab} &  Motion & \phantom{ab} & App.+Motion\\
\hline
Improved Dense Trajectories~\cite{wang:hal-00873267} & - & & - & & 0.141\\
Dynamic Image Networks~\cite{Bilen16a}  & 0.231 & & - & & - \\
Two Stream Architecture~\cite{NIPS2014_5353}  & 0.270 & & 0.163 & & 0.120\\
Two Stream Architecture (GoogleNet)~\cite{DBLP:journals/corr/WangXW015}  & 0.247 & & 0.142 & & 0.107\\
Two Stream Architecture (VGG-Net)~\cite{DBLP:journals/corr/WangXW015} & 0.216 & & 0.130 & & 0.086\\
Spatiotemporal ResNets~\cite{NIPS2016_6433} & - & & - & & 0.066\\
\hline
Recurrent Residual Networks & \textbf{0.198} & & - & & -\\
\hline
\end{tabularx}
\caption{Classification error rates for UCF-101. }
\label{table_compare_state_of_the_art}
\vspace{-0.7cm}
\end{table}
Our model achieves better error rates than the state of the art appearance stream models. Only fused models perform better. Note that the works~\cite{NIPS2014_5353,DBLP:journals/corr/WangXW015,NIPS2016_6433} are all two-stream architectures. The dynamic image network~\cite{Bilen16a} is a purely appearance base method that reduces that video to a single frame and uses a ConvNet to classify this frame. For a fair comparison, we provide the result of dynamic image network without the combination with dense trajectories as this would include motion features.

\section{Conclusion}
\vspace{-0.2cm}

We extended the ResNet architecture to include temporal skip connections in order to model both, spatial and temporal information in video. Our model performs well already with a single temporal skip connection, enabling to infer context between two frames. Moreover, we showed that fusing temporal information at a late stage in the network is beneficial and that learning a temporal residual is superior to using a classical recurrent layer. Our method is not limited to appearance based models and can easily be extended to motion networks that have shown to further enhance the performance on action recognition datasets. A comparison to both, a ResNet baseline and state of the art methods showed that our approach outperforms other purely appearance based approaches.\\
\par
\textbf{Acknowledgement:} The authors have been financially supported by the DFG projects KU 3396/2-1 and GA 1927/4-1 and the ERC Starting Grant ARCA (677650). Further, this work was supported by the AWS Cloud Credits for Research program.

	\bibliographystyle{splncs03}
	\bibliography{egbib}

\end{document}